\documentclass[letterpaper,journal]{IEEEtran}
\usepackage{amsmath,amsfonts}
\usepackage{algorithmic}
\usepackage{algorithm}
\usepackage{array}
\usepackage[caption=false,font=normalsize,labelfont=sf,textfont=sf]{subfig}
\usepackage{textcomp}
\usepackage{stfloats}
\usepackage{multirow}
\usepackage{url}
\usepackage{verbatim}
\usepackage{booktabs} 
\usepackage{graphicx}
\usepackage{cite}
\usepackage{xcolor}
\begin{document}

\title{AdNGCL: Adaptive Negative Scheduling for Graph Contrastive Learning}

\author{Adnan Ali,
        Jinlong Li,
        Syed Muhammad Israr,
        Guanghua Yang \IEEEmembership{Senior Member,~IEEE},
        Ali Kashif Bashir \IEEEmembership{Senior Member,~IEEE},
        and Mohammed Ameen

\thanks{This work has been submitted to the IEEE for possible publication. Copyright may be transferred without notice, after which this version may no longer be accessible.}
       
\thanks{Adnan Ali (Corresponding author) and Guanghua Yang are with the School of Intelligent Systems Science and Engineering, Jinan University, Zhuhai 519070, China (e-mails: adnanali@jnu.edu.cn; ghyang@jnu.edu.cn).}

\thanks{Jinlong Li is with the School of Computer Science and Technology, University of Science and Technology of China, Hefei, China (e-mail: jlli@ustc.edu.cn).}

\thanks{Syed Muhammad Israr is with Hainan University, Haikou, China (e-mail: syed\_ustc@hainanu.edu.cn).}

\thanks{Ali Kashif Bashir is with the Department of Computing and Mathematics, Manchester Metropolitan University, Manchester, United Kingdom (e-mail: dr.alikashif.b@ieee.org).}

\thanks{Mohammed Ameen is with the Department of Information Systems, King Abdulaziz University, Rabigh, Saudi Arabia (e-mail: mkameen@kau.edu.sa).}
}

\maketitle

\begin{abstract}
Negative samples play a crucial role in graph contrastive learning (GCL), yet existing methods mainly focus on which negatives should be selected, mined, filtered, or progressively introduced. Less attention has been given to separating the negative candidates available to an anchor from the subset that actively participates in optimization, and to adapting this exposure according to training behavior.
To address this gap, we propose Adaptive Negative Scheduling for Graph Contrastive Learning (AdNGCL) framework, which maintains a bounded per-anchor candidate reservoir and periodically reassesses candidate hardness as representations evolve.
Its core component, Hardness-Aware Negative Scheduling (HANS), uses a category-specific training-response signal to determine whether exposure to easy, intermediate, or hard negatives should increase at each scheduling opportunity. 
Furthermore, reliability-aware candidate refresh handles potentially positive-like candidates, while within-category diversity refresh changes active identities without altering HANS-controlled exposure. 
AdNGCL retains a standard symmetric InfoNCE objective. 
Experiments on ten benchmark datasets against state-of-the-art self-supervised graph learning methods show that AdNGCL achieves the highest mean Micro-F1 among the evaluated methods on eight datasets, with improvements of up to \textit{1.30 percentage points} over the second-best method, ranks second on Actor, and remains competitive on Coauthor-CS. Ablation, sensitivity, scheduler and reliability diagnostics, and computational analyses further examine the behavior and contribution of the proposed mechanisms.
\end{abstract}

\begin{IEEEkeywords}
Graph contrastive learning, self-supervised learning, negative sampling, adaptive negative exposure, hardness-aware scheduling, candidate reliability.
\end{IEEEkeywords}

\section{Introduction}

\IEEEPARstart{G}{raph}-structured data are widely used in applications such as adversarially robust learning~\cite{adv10494404,ad11456296}, cross-domain transfer learning~\cite{9739062}, and heterogeneous information networks~\cite{10296092}. However, learning useful representations from such data remains challenging because graph observations combine node attributes with irregular relational structure that cannot be handled directly by conventional Euclidean learning models~\cite{10382709}. Graph Neural Networks provide an effective mechanism for exploiting this structure and have achieved strong performance across node, edge, and graph-level prediction tasks~\cite{JU2024106207}. Their effectiveness, however, often depends on labeled supervision, which can be costly or unavailable in many graph applications~\cite{9737635}. This limitation has motivated self-supervised graph representation learning, particularly Graph Contrastive Learning (GCL), where representations are learned from the graph itself without requiring class labels~\cite{11084849}.

\par
A typical GCL framework constructs augmented graph views through stochastic structural and feature augmentations and trains an encoder to increase agreement between representations of the same underlying instance (positive sample) while distinguishing them from negative samples~\cite{MHGCLHA}. 
Earlier methods ~\cite{DeepGrace2020,GCADBLP-abs-2010-14945} focused on data augmentation and have demonstrated the effectiveness of this learning paradigm for graph representation learning. 
However, negative sampling has received comparatively less attention. 
The positive pairs define the representations that should remain consistent across augmented views, whereas \textit{negative samples} provide the contrasting instances from which an anchor representation should be separated~\cite{dataaug11194909}.
Consequently, the composition and use of negative samples directly influence the contrastive discrimination imposed on the encoder, making negative management an important component of the GCL training process~\cite{NegAmplify}. 
Uniform negative sampling, however, treats all candidate negatives alike despite substantial differences in their informativeness and difficulty, which can lead to large numbers of weak or redundant contrasts and inefficient use of the negative set and computational resources.

\par
Motivated by these limitations, a growing body of work has moved beyond uniform negative sampling toward more informative negative-management strategies.
ProGCL~\cite{xia2022progcl} progressively incorporates informative negatives while accounting for the uncertainty of hard negative identification. NegAmplify~\cite{NegAmplify} explicitly considers the quantity, quality, and hardness composition of negative samples. More recent approaches examine complementary aspects of the same problem: E2Neg~\cite{E2Neg} seeks a compact set of effective and efficient negatives, whereas IFL-GCL~\cite{IFLGCL} addresses false-negative contamination using semantic information derived from the contrastive process. These developments show that negative management has evolved beyond simple random sampling toward decisions about informativeness, hardness, compactness, and reliability.

\par

Despite these advances, existing methods still largely formulate negative management in terms of selecting, mining, filtering, or ordering candidate negatives~\cite{cucoijcai,HardNegativerobinson2021}.
Studies of negative sampling have shown that the usefulness of a negative depends on factors such as hardness, false-negative risk, redundancy, and the stage of representation learning, while graph-specific investigations further demonstrate that increasing the number of negatives does not necessarily provide proportional benefit~\cite{NegAmplify,E2Neg,hal-03575619}. Curriculum-based strategies partially account for this temporal variation by changing which negatives are introduced during training~\cite{cucoijcai,ConCur}. However, a distinct question remains: how should a bounded set of available negative candidates be separated from the subset that actively participates in contrastive optimization, and how should that exposure change in response to the observed training behavior? This motivates treating negative exposure as an adaptive control problem alongside negative selection itself.

\par

To address this problem, we propose \emph{Adaptive Negative Scheduling for Graph Contrastive Learning} (AdNGCL). \emph{First}, AdNGCL maintains a bounded set of available negative candidates and periodically reassesses their relative hardness as the representation space evolves. \emph{Second}, its central component, \emph{Hardness-Aware Negative Scheduling (HANS)}, controls how many easy, intermediate, and hard negatives actively participate in optimization according to recent training response rather than increasing exposure automatically at predetermined training intervals.
\emph{Third}, reliability-aware candidate refresh handles potentially positive-like reservoir candidates, while within-category refresh changes active candidate identities without altering the exposure determined by HANS. In this way, AdNGCL separates negative availability, hardness organization, and training-time exposure within a unified contrastive framework.

\par

The key distinction of AdNGCL lies in treating negative \emph{exposure} as a separate adaptive decision rather than only determining which negatives should be selected. ProGCL~\cite{xia2022progcl} progressively incorporates informative negatives, NegAmplify~\cite{NegAmplify} considers negative quantity, quality, and hardness composition, E2Neg~\cite{E2Neg} seeks compact and efficient negative sets, and IFL-GCL~\cite{IFLGCL} focuses on mitigating false negatives. AdNGCL complements these directions by maintaining a bounded candidate space while independently controlling when and how much of that space participates in optimization according to the observed training response. Thus, its primary novelty is response-driven negative exposure, with dynamic hardness reassessment and reliability/diversity maintenance supporting that scheduling process.
The main contributions of this work are summarized as follows:

\begin{itemize}

    \item We formulate negative management in GCL by separating the negative candidates \emph{available} to an anchor from the subset \emph{actively exposed} during optimization. Based on this formulation, we introduce Hardness-Aware Negative Scheduling (HANS), which uses recent category-specific training response to control the timing and amount of negative exposure rather than increasing negative exposure solely according to a fixed epoch-driven schedule.

    \item We develop an adaptive candidate-management framework around HANS. Candidate hardness is periodically reassessed as the learned representation changes, while a conservative reliability-aware refresh handles positive-like candidates under bidirectional and hardness-preservation constraints. A within-category refresh further changes active negative identities without modifying the exposure selected by the scheduler.

    \item We evaluate AdNGCL on ten graph benchmark datasets against representative state-of-the-art self-supervised graph learning methods, including ProGCL, NegAmplify, E2Neg, IFL-GCL, and MHGCL-HA. AdNGCL achieves the highest mean Micro-F1 on eight of ten datasets, with gains of up to 1.30 percentage points over the second-best method, ranks second on Actor, and remains competitive on Coauthor-CS. Component ablations, parameter sensitivity, scheduler and reliability diagnostics, and controlled computational analysis further evaluate the proposed mechanisms.

\end{itemize}

The remainder of this paper is organized as follows. Section~\ref{sec:relatedWork} reviews graph contrastive learning and negative-management strategies. Section~\ref{sec:method} presents the proposed AdNGCL framework. Section~\ref{sec:experiments} reports the experimental setup, overall performance, sensitivity analysis, component analysis, scheduler and reliability diagnostics, and computational evaluation. Finally, Section~\ref{sec:conclusion} concludes the paper.

\section{Related Works}
\label{sec:relatedWork}
This section reviews representative developments in graph contrastive learning and prior work on negative sample management, including hardness-aware, reliability-aware, compact, and curriculum-based strategies 
\subsection{Graph Contrastive Learning}

Graph Contrastive Learning (GCL) has become an important direction in self-supervised graph representation learning because it enables task-independent representations to be learned without annotations~\cite{9739062}. Early approaches such as DGI~\cite{velikovi2019deep} maximized mutual information between local node representations and a global graph summary, while MVGRL~\cite{pmlr-v119-hassani20a} extended this idea through multiple structural views based on graph adjacency and diffusion. Later, GRACE~\cite{DeepGrace2020} introduced an augmentation-based node-level contrastive framework using stochastic edge dropping and feature masking, and GCA~\cite{GCADBLP-abs-2010-14945} further adapted these augmentations according to graph topology. These developments established multi-view augmentation and contrastive agreement as effective mechanisms for self-supervised graph representation learning.

\par
Graph self-supervised learning has subsequently expanded beyond conventional negative-based contrastive objectives. BGRL~\cite{thakoor2021bootstrapped} adopts a bootstrap learning strategy to align representations across augmented views without explicit negative samples, while Graph Barlow Twins~\cite{BIELAK2022109631} learns representations through redundancy reduction. Augmentation-free approaches, including AFGRL~\cite{AFGRL} and AFGCL~\cite{AFGCL}, further reduce reliance on manually designed graph perturbations. Recent methods have also continued to improve representation learning through architectural, augmentation, and multi-level contrastive mechanisms~\cite{MHGCLHA}. Together, these studies show that the effectiveness of graph self-supervised learning depends on several interacting design choices, including view construction, representation alignment, and the contrastive objective.

\par
For GCL methods that retain an explicit contrastive objective, the construction and use of negative samples remain an important part of the learning process. Negative candidates can differ substantially in their informativeness, similarity to the anchor, redundancy, and likelihood of representing semantically related instances~\cite{9770382}. Consequently, treating all available candidates in the same manner may provide limited or unreliable contrastive information \cite{NegAmplify}. This motivated how negative samples should be selected, evaluated, and incorporated during graph contrastive training.

\subsection{Negative Sample Management in GCL}
Negative sampling directly determines the competing instances used by contrastive objectives and therefore has an important effect on the supervision received by the encoder~\cite{HardNegativerobinson2021,negativeReview}. Early GCL methods commonly relied on in-batch or uniformly sampled negatives, where non-positive instances were treated similarly during optimization~\cite{DeepGrace2020,GCADBLP-abs-2010-14945,zhu2021empirical}. Although simple and scalable, such strategies do not distinguish between informative and weak negatives and may introduce substantial redundancy as the negative set grows. Graph-specific investigations have further shown that increasing the number of negatives does not necessarily yield proportional improvements in representation quality~\cite{hal-03575619}. These observations motivated more selective negative-management strategies.

\par
A major direction focuses on negative \emph{hardness} and informativeness. Hard-negative mining emphasizes candidates that are more similar to the anchor and therefore impose a more difficult discrimination task~\cite{HardNegativerobinson2021,CLHNS2020_f7cade80}. In GCL, ProGCL~\cite{xia2022progcl} estimates the probability that hard candidates are true negatives and progressively incorporates informative samples, while NegAmplify~\cite{NegAmplify} considers negative quantity, quality, and hardness composition. AUGCL~\cite{10382709} uses uncertainty to estimate negative importance, and AdaS~\cite{10181235} adaptively models negative importance while accounting for potential false negatives. Collectively, these methods demonstrate that negative candidates should not be treated uniformly and that their difficulty and informativeness can substantially affect contrastive learning.

\par
Negative hardness, however, is closely related to reliability because highly similar candidates may correspond to semantically related instances rather than valid negatives. This has motivated methods that explicitly address false-negative contamination and candidate quality. IFL-GCL~\cite{IFLGCL}, for example, identifies potentially unreliable negatives using semantic information derived during contrastive learning, while neighborhood- and structure-aware strategies exploit graph relations to reduce the likelihood of treating related nodes as negatives~\cite{Zhang2024}. A complementary line of work considers the size and efficiency of the negative set. E2Neg~\cite{E2Neg} seeks a compact collection of effective negatives, reflecting the observation that a larger candidate set may contain substantial redundancy and does not necessarily provide proportional training benefit.

\par
Other studies consider how negative samples should change over the course of training. Curriculum-based approaches such as CuCo~\cite{cucoijcai} organize negative samples according to difficulty and introduce them progressively, while ConCur~\cite{ConCur} further explores curriculum-based negative sampling in graph representation learning. ProGCL~\cite{xia2022progcl} similarly reflects the importance of evolving negative selection as learned representations become more reliable. These methods establish that negative management need not remain fixed throughout optimization and that the usefulness of a candidate can depend on the current stage of representation learning.

\par
Overall, existing research has addressed negative informativeness, hardness, reliability, compactness, and progressive incorporation from different perspectives. These approaches primarily determine which candidates should be selected, filtered, ranked, or introduced during training. A less explored aspect is the explicit separation between a bounded set of negatives that are available to an anchor and the subset that actively participates in the contrastive objective, together with adapting that exposure according to the observed training response. This distinction motivates the negative-exposure formulation developed in the following section.

\section{Proposed AdNGCL}
\label{sec:method}
\begin{figure*}[!t]
\centering
\includegraphics[width=1\textwidth]{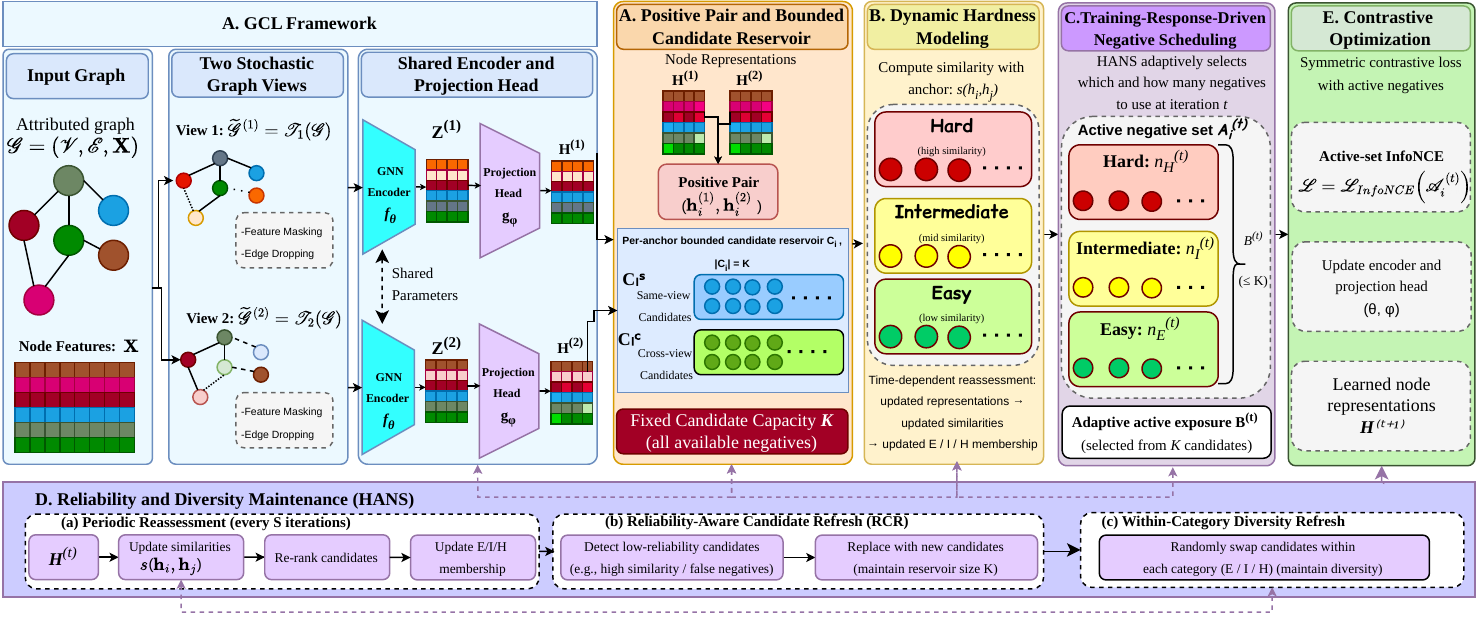}
\caption{Overview of the proposed AdNGCL framework. Two stochastic graph views are processed by a shared encoder and projection head to construct positive pairs and a bounded reservoir of negative candidates with fixed capacity $K$. Candidate hardness is dynamically reassessed according to anchor-candidate similarity and organized into easy, intermediate, and hard groups. HANS adaptively controls the active negative exposure $B^{(t)}\leq K$ using category-specific training-response signals, while reliability-aware candidate refresh and within-category diversity refresh support candidate reliability and diversity. Only the currently active negatives participate in the symmetric active-set InfoNCE objective for representation learning.}
\label{AdNGCLModel}
\end{figure*}

AdNGCL is a graph contrastive learning framework that adaptively controls negative-sample exposure during training. It distinguishes between negatives that are \emph{available} to an anchor and those that are \emph{active} in the contrastive objective. A bounded candidate reservoir defines the available space, while Hardness-Aware Negative Scheduling (HANS) regulates active exposure according to training response. The overall framework is illustrated in Fig.~\ref{AdNGCLModel}.

The subsection \textit{Framework and Bounded Candidate Reservoir} defines the graph contrastive learning setting and introduces the distinction between bounded candidate capacity and temporally active negative exposure. \textit{Dynamic Hardness Modeling} describes how candidates are organized into relative easy, intermediate, and hard categories and periodically reassessed as the representation space evolves. \textit{Training-Response-Driven Negative Scheduling} introduces HANS, which uses category-specific training signals to determine when and how much additional negative exposure is introduced. \textit{Reliability and Diversity Maintenance} presents the reliability-aware candidate refresh and within-category diversity refresh mechanisms used to maintain candidate reliability and diversity. Finally, \textit{Contrastive Optimization and Complexity} specifies the active-set InfoNCE objective and analyzes the computational and memory requirements of AdNGCL.

\subsection{Framework and Bounded Candidate Reservoir}
\label{subsec:framework}
This section presents the basic graph contrastive learning formulation of AdNGCL and introduces the bounded candidate reservoir that separates the available negative capacity from the subset actively used during optimization.
\par
\textit{First}, the basic graph contrastive learning formulation is presented as: let an attributed graph be $\mathcal{G}=(\mathcal{V},\mathcal{E},\mathbf{X})$,
where $|\mathcal{V}|=N$ and $\mathbf{X}\in\mathbb{R}^{N\times F}$. 
We generate two stochastically augmented views by applying graph augmentation operators $\mathcal{T}_1$ and $\mathcal{T}_2$, consisting of feature masking and edge dropping:
$\widetilde{\mathcal{G}}^{(1)}=\mathcal{T}_1(\mathcal{G})$ and
$\widetilde{\mathcal{G}}^{(2)}=\mathcal{T}_2(\mathcal{G})$.
Furthermore, a shared graph encoder $f_{\theta}$ and projection head $g_{\phi}$ are defined as
$\mathbf{Z}^{(r)}=f_{\theta}(\widetilde{\mathcal{G}}^{(r)})$ and
$\mathbf{H}^{(r)}=g_{\phi}(\mathbf{Z}^{(r)})$,
where $r\in\{1,2\}$. For node $v_i\in\mathcal{V}$, $\mathbf{h}^{(1)}_i$ and $\mathbf{h}^{(2)}_i$ form its positive pair.
\par

\textit{Second}, we introduce the bounded candidate reservoir used to define the negative search space of AdNGCL. For each anchor node $v_i$, negative candidates are drawn from nodes $v_j$ with $j\neq i$, thereby excluding the anchor identity from the negative set. Each candidate position is assigned either a same-view or a cross-view role according to the representation used in the corresponding contrastive direction.
\par
Rather than exposing the anchor to all admissible negatives, AdNGCL maintains a fixed-capacity reservoir of $K$ candidate positions,
\begin{equation}
\mathcal{C}_i=\mathcal{C}^{s}_i\cup\mathcal{C}^{c}_i,\qquad|\mathcal{C}_i|=K,
\label{eq:candidate_reservoir}
\end{equation}
where $\mathcal{C}^{s}_i$ and $\mathcal{C}^{c}_i$ denote the same-view and cross-view candidate subsets, respectively. Candidate identities are unique within $\mathcal{C}_i$, and the implementation allocates an equal number of reservoir positions to the two roles. The capacity $K$ therefore fixes the maximum negative candidate space available to each anchor, independently of how many of those candidates are active in the contrastive objective at a given training stage.
\par
The bounded reservoir defines the negative candidates available to an anchor, but only a subset of these candidates is used in the contrastive objective at a given training stage. We denote the corresponding active negative count at epoch $t$ by
\begin{equation}
    B^{(t)} \leq K.
\end{equation}
Thus, $K$ remains the fixed candidate capacity, whereas $B^{(t)}$ determines the amount of negative exposure during training. The evolution of $B^{(t)}$ is controlled by HANS, which is introduced in Section~\ref{subsec:hans}.

\subsection{Dynamic Hardness Modeling}
\label{subsec:hardness}
After constructing the bounded reservoir of $K$ negative candidates for each anchor, AdNGCL next organizes these candidates according to their similarity to the anchor. This section defines the resulting relative hardness measure and explains how the candidates are assigned to easy, intermediate, and hard categories during training.
\par
We define negative hardness relative to the current projected representation of each anchor. For projected representations $\mathbf{h}_i$ and $\mathbf{h}_j$, cosine similarity is
\begin{equation}
s(\mathbf{h}_i,\mathbf{h}_j)
=
\frac{\mathbf{h}_i^{\top}\mathbf{h}_j}
{\|\mathbf{h}_i\|_2\|\mathbf{h}_j\|_2}.
\label{eq:cosine}
\end{equation}

For each contrastive direction, the $K$ candidate positions of an anchor are ranked by their current similarity to that anchor and divided into approximately equal rank-based thirds,
\begin{equation}
\mathcal{C}^{(d)}_{i}
=
\mathcal{C}^{E,(d)}_{i}
\cup
\mathcal{C}^{I,(d)}_{i}
\cup
\mathcal{C}^{H,(d)}_{i},
\qquad
d\in\{1{\rightarrow}2,2{\rightarrow}1\},
\label{eq:hardness_partition}
\end{equation}
where lower similarities form the easy category, intermediate similarities form the intermediate category, and larger similarities form the hard category. These categories represent \emph{relative hardness within the anchor-specific reservoir}; they are not semantic classes.

Since hardness is computed from the learned representations, it is inherently time-dependent. Specifically, the similarity $s_{ij}^{(t)}$ and therefore the rank of candidate $v_j$ within the reservoir of anchor $v_i$ may change as the encoder parameters are updated. AdNGCL consequently recomputes the directional partitions $\mathcal{C}_{i}^{E,(d,t)}$, $\mathcal{C}_{i}^{I,(d,t)}$, and $\mathcal{C}_{i}^{H,(d,t)}$ from the current projected representations at periodic reassessment epochs.

Following each reassessment, the active sets are reconciled with the updated partitions such that
\[
\mathcal{A}_{i}^{c,(t)}\subseteq\mathcal{C}_{i}^{c,(d,t)},
\qquad
|\mathcal{A}_{i}^{c,(t)}|=B_c^{(t)},
\]
for $c\in\{E,I,H\}$. Existing active candidates are retained when they remain in the corresponding updated category; otherwise, the required positions are sampled from that category. This establishes a separation between the two adaptive components of AdNGCL: dynamic hardness modeling updates the \emph{membership} of the E/I/H candidate sets, whereas HANS updates their active \emph{cardinalities} through $B_c^{(t)}$.

Hardness reassessment is performed periodically using the same temporal cadence as the HANS scheduling process. The reassessment updates the E/I/H candidate memberships throughout training, whereas changes to the active category counts $B_c^{(t)}$ are governed separately by the HANS adaptation rule.

\subsection{Training-Response-Driven Negative Scheduling}
\label{subsec:hans}
Having organized the bounded candidate reservoir into dynamic easy, intermediate, and hard categories, AdNGCL next determines how much of each category participates in contrastive optimization at each stage of training. To achieve this, we introduce \textit{Hardness-Aware Negative Scheduling (HANS)}, a training-response-driven mechanism that maintains and updates the active negative count of each hardness category over time. HANS therefore controls both the amount and timing of negative exposure, while candidate reliability and diversity are addressed separately in the following subsection.
\par
For each hardness category $c\in\{E,I,H\}$, HANS maintains a utilization variable
$\eta_c^{(t)}\in[0,1]$. If $K_c$ denotes the capacity of category $c$, the corresponding active count is
\begin{equation}
B_c^{(t)}
=
\operatorname{round}\!\left(\eta_c^{(t)}K_c\right),
\qquad
0\leq B_c^{(t)}\leq K_c .
\label{eq:active_count}
\end{equation}
For anchor $i$,
$\mathcal{A}^{c,(t)}_i\subseteq\mathcal{C}^{c,(t)}_i$
denotes the active subset of category $c$, and
\begin{equation}
\mathcal{A}^{(t)}_i
=
\mathcal{A}^{E,(t)}_i
\cup
\mathcal{A}^{I,(t)}_i
\cup
\mathcal{A}^{H,(t)}_i .
\end{equation}
All categories begin from the same small utilization $\eta_0$. Their utilizations are non-decreasing, but need not evolve identically because each category is gated independently by its observed training response. Active candidates are sampled from the corresponding hardness category and existing active candidates are retained whenever possible as the count grows.

\paragraph{Category-specific training-response signal.}
A raw sum over negative contributions grows directly with the number of active negatives and is therefore unsuitable for comparing scheduler behavior across changing exposure levels. HANS instead uses a count-normalized contrastive difficulty signal. For one direction, anchor $i$, and category $c$ with $B_c^{(t)}>0$,
\begin{equation}
D_{i,c}^{(t)}
=
\log
\left[
1+
\frac{1}{B_c^{(t)}}
\sum_{j\in\mathcal{A}^{c,(t)}_i}
\exp
\left(
\frac{s_{ij}^{-,(t)}-s_i^{+,(t)}}{\tau}
\right)
\right],
\label{eq:category_signal_anchor}
\end{equation}
where $s_i^{+}$ denotes positive-pair similarity, $s_{ij}^{-}$ denotes active-negative similarity, and $\tau$ is the contrastive temperature. When a category contains no active candidate, its scheduler signal is defined as zero.

Signals are computed independently in both contrastive directions and averaged:
\begin{equation}
D_c^{(t)}
=
\frac{1}{2N}
\sum_{i=1}^{N}
\left(
D_{i,c}^{1\rightarrow2,(t)}
+
D_{i,c}^{2\rightarrow1,(t)}
\right).
\label{eq:category_signal}
\end{equation}

The difference $s_{ij}^{-}-s_i^{+}$ represents negative-positive competition: the signal increases when active negatives become more competitive with the positive pair. Dividing by $B_c^{(t)}$ reduces the direct dependence of signal magnitude on the number of active negatives, making the signal more comparable as exposure changes. Importantly, $D_c^{(t)}$ is detached from gradient computation and is used only by the scheduler; it is not an additional representation-learning loss.

\paragraph{Training-response-gated exposure.}
After an initial bootstrap period, HANS considers one hardness category at each scheduling opportunity in the rotating order
\[
H\rightarrow I\rightarrow E .
\]
Let $W$ denote the signal-window length. Two consecutive windows are summarized as
\begin{equation}
D_{c,\mathrm{prev}}
=
\sum_{q=t-2W+1}^{t-W}D_c^{(q)},
\qquad
D_{c,\mathrm{cur}}
=
\sum_{q=t-W+1}^{t}D_c^{(q)} .
\label{eq:signal_windows}
\end{equation}
Additional exposure is permitted when
\begin{equation}
D_{c,\mathrm{cur}}
\ge
\gamma D_{c,\mathrm{prev}},
\label{eq:hans_gate}
\end{equation}
where $\gamma\in(0,1]$ controls the required change in the recent training-response signal. For $\gamma=0.99$, for example, the gate passes when category difficulty has decreased by less than approximately $1\%$ across the two windows, or has increased. Thus, additional negatives are exposed when progress under the current active set has slowed rather than according to epoch count alone.

When the gate passes,
\begin{equation}
\eta_c^{(t+1)}
=
\min\!\left(1,\eta_c^{(t)}+\Delta\eta\right),
\qquad
\eta_c^{(t+1)}=\eta_c^{(t)}
\quad\text{otherwise}.
\label{eq:eta_update}
\end{equation}
Rather than treating the utilization increment as an additional tunable parameter, AdNGCL derives
\begin{equation}
\Delta\eta
=
\frac{1-\eta_0}{Q},
\label{eq:eta_step}
\end{equation}
where $Q$ is the number of scheduling opportunities assigned to each hardness category during the adaptive scheduling interval. Consequently, a category can reach full utilization if all of its gates pass, whereas failed gates leave that category only partially exposed.

Under the shared experimental configuration, training lasts $T=1000$ epochs, the first $20\%$ constitutes the bootstrap stage, and HANS is permitted to adapt utilization until $80\%$ of training. With $W=10$ and scheduling opportunities every $2W=20$ epochs, the adaptive interval contains $30$ scheduler events and $Q=10$ opportunities per hardness category. With $\eta_0=0.05$, this yields $\Delta\eta=0.095$. These values are shared across datasets rather than tuned independently.

\subsection{Reliability and Diversity Maintenance}
\label{subsec:maintenance}

HANS controls the number and timing of active negatives from each hardness category, but hardness alone does not establish whether a highly similar candidate is a reliable negative. We therefore complement the exposure scheduler with two candidate-maintenance operations. Reliability-aware Candidate Refresh (RCR) examines potentially positive-like candidates in the bounded reservoir, while within-category refresh periodically changes the identities of active negatives without altering their hardness category or active count.

\paragraph{Reliability-aware candidate refresh.}
For a candidate $v_j$ associated with anchor $v_i$, let $s_{ij}^{-,12}$ and $s_{ij}^{-,21}$ denote its negative similarities in the two contrastive directions. A candidate is marked as \emph{positive-like} only when its similarity reaches or exceeds the positive-pair similarity in both directions:
\begin{equation}
s_{ij}^{-,12}\ge s_i^{+}
\quad\land\quad
s_{ij}^{-,21}\ge s_i^{+}.
\label{eq:suspicious}
\end{equation}
The use of both directions restricts replacement to candidates for which the positive-like behavior is consistent across the two views. No class labels are used in this decision.

When RCR is performed, a temporary set of $K$ non-anchor proposals is sampled for each anchor. This proposal set is used only for candidate replacement and does not modify the persistent reservoir capacity $K$ or the active count $B^{(t)}$. For a positive-like candidate $\bar{j}$, a proposal $j$ is considered eligible only if: (i) it is not already contained in the current reservoir; (ii) its similarity remains below the positive-pair similarity in both contrastive directions; (iii) it belongs to the same pre-refresh E/I/H rank category as $\bar{j}$ in both directions; and (iv) it has not already been selected as another replacement for the same anchor.

Let $\mathcal{R}_i(\bar{j})$ denote the proposals satisfying these constraints. Among them, RCR selects the candidate with the smallest change in directional hardness,
\begin{equation}
j^{*}
=
\arg\min_{j\in\mathcal{R}_i(\bar{j})}
\left[
\left(s_{ij}^{-,12}-s_{i\bar{j}}^{-,12}\right)^2
+
\left(s_{ij}^{-,21}-s_{i\bar{j}}^{-,21}\right)^2
\right].
\label{eq:rcr_selection}
\end{equation}
If $\mathcal{R}_i(\bar{j})$ is empty, the original candidate is retained. Replacement also preserves the same-view or cross-view role of the corresponding reservoir position. Hence, RCR changes candidate identity only when an eligible hardness-preserving replacement is available, while constraining the replacement to remain in the same hardness regime.

RCR is coupled with dynamic hardness reassessment. At a reassessment event, reliability checking is first applied to the current reservoir, after which the directional E/I/H partitions are recomputed. The updated hardness model therefore operates on the refreshed candidate reservoir.

\paragraph{Within-category diversity refresh.}
RCR modifies the reservoir only when a candidate satisfies the positive-like criterion. A separate refresh operation is used to prevent the active set from repeatedly containing the same candidate identities. For each category $c$, a fraction of the currently active positions is exchanged with inactive positions from the same current hardness pool. If $\mathcal{A}^{c}_i$ denotes the active set before the refresh, the resulting set $\widetilde{\mathcal{A}}^{c}_i$ satisfies
\begin{equation}
\widetilde{\mathcal{A}}^{c}_i
\subseteq
\mathcal{C}^{c}_i,
\qquad
|\widetilde{\mathcal{A}}^{c}_i|
=
|\mathcal{A}^{c}_i|.
\label{eq:diversity_refresh}
\end{equation}
The operation therefore changes negative identities without changing either the hardness category or the active count selected by HANS.

The three mechanisms act on different quantities. Dynamic hardness reassessment updates the \emph{category membership} of reservoir candidates, RCR updates \emph{candidate identities} when positive-like behavior is detected, and within-category refresh updates the \emph{active identities} while preserving the HANS-controlled exposure.

\subsection{Contrastive Optimization and Computational Complexity}
\label{subsec:objective}

After HANS determines the active negative exposure and the maintenance mechanisms update the candidate reservoir, representation learning is performed using only the resulting active negatives. AdNGCL does not introduce a new contrastive loss; instead, it changes which negatives participate in the standard objective and when they are introduced during training.

For anchor $v_i$ in the $1\rightarrow2$ direction, the active-set InfoNCE loss is
\begin{equation}
\mathcal{L}^{1\rightarrow2}_i
=
-\log
\frac{
\exp(s_i^{+,1\rightarrow2}/\tau)
}{
\exp(s_i^{+,1\rightarrow2}/\tau)
+
\sum_{j\in\mathcal{A}^{1\rightarrow2,(t)}_i}
\exp(s_{ij}^{-,1\rightarrow2}/\tau)
}.
\label{eq:infonce}
\end{equation}
The same objective is evaluated in the reverse direction, giving the symmetric training objective
\begin{equation}
\mathcal{L}_{\mathrm{AdNGCL}}
=
\frac{1}{2N}
\sum_{i=1}^{N}
\left(
\mathcal{L}^{1\rightarrow2}_i
+
\mathcal{L}^{2\rightarrow1}_i
\right).
\label{eq:final_loss}
\end{equation}
All active negatives contribute to the denominator in the same form; no hardness-dependent weighting is introduced. Hence, the role of AdNGCL is to control the membership, cardinality, and timing of the active negative set, while the encoder and projection head are optimized through the standard symmetric InfoNCE objective.

\paragraph{Effect of Negative Exposure.}
The active-negative count and the relative difficulty of the exposed candidates jointly determine the InfoNCE objective. A supporting theoretical analysis, provided in Supplementary Section S1, derives the exact change in the contrastive loss under active-set expansion and establishes a conditional bound on the corresponding gradient perturbation. This analysis characterizes the immediate optimization effect of changing negative exposure without asserting a general convergence guarantee.

\paragraph{Computational complexity.}
Let
\begin{equation}
B^{(t)}
=
B_E^{(t)}+B_I^{(t)}+B_H^{(t)}
\leq K
\end{equation}
denote the total number of active negatives per anchor at epoch $t$. Excluding the graph-encoder computation shared with the underlying GCL framework, evaluating the active negative similarities requires
\begin{equation}
\mathcal{O}\!\left(NB^{(t)}d\right),
\end{equation}
where $d$ is the projected representation dimension. The category-specific HANS signals are computed from the same active candidates and have the same asymptotic dependence on $N$, $B^{(t)}$, and $d$. When the complete reservoir is active, the corresponding worst-case negative-comparison cost becomes $\mathcal{O}(NKd)$.

Periodic candidate maintenance introduces additional computation. Recomputing the similarities of the $K$ reservoir candidates and ranking them into the E/I/H categories requires
\begin{equation}
\mathcal{O}\!\left(NKd+NK\log K\right).
\end{equation}
For RCR, a temporary set of $K$ replacement proposals is additionally evaluated for each anchor. Computing their directional similarities requires $\mathcal{O}(NKd)$ operations, while checking eligible proposals and selecting hardness-preserving replacements has a worst-case $\mathcal{O}(NK^2)$ cost. A complete reliability and hardness refresh therefore has worst-case complexity
\begin{equation}
\mathcal{O}\!\left(
NKd+NK\log K+NK^2
\right),
\label{eq:refresh_complexity}
\end{equation}
but this cost is incurred only at periodic reassessment events rather than at every optimization epoch.

The persistent candidate reservoir requires $\mathcal{O}(NK)$ index storage, while the node representations require $\mathcal{O}(Nd)$ memory. The temporary RCR proposal set also requires $\mathcal{O}(NK)$ indices but does not increase the persistent reservoir capacity. Therefore, for a fixed candidate capacity $K$, the negative-management cost remains linear in the number of graph nodes, while $K$ controls the trade-off between the size of the candidate search space and its computational and memory requirements.

Algorithm~\ref{alg:adngcl} summarizes the complete AdNGCL training procedure.

\begin{algorithm}[t]
\caption{Adaptive Negative Scheduling for Graph Contrastive Learning (AdNGCL)}
\label{alg:adngcl}
\begin{algorithmic}[1]

\REQUIRE Graph $\mathcal{G}$; encoder $f_{\theta}$; projector $g_{\phi}$;
$K,\eta_0,W,\gamma,Q,T$
\ENSURE Trained encoder $f_{\theta}$

\STATE $T_b\!\leftarrow\!0.2T$, $T_a\!\leftarrow\!0.8T$,
$\Delta\eta\!\leftarrow\!(1-\eta_0)/Q$
\STATE Construct $K$-candidate reservoirs and directional E/I/H pools
\STATE $\eta_E=\eta_I=\eta_H=\eta_0$; initialize active sets

\FOR{$t=1,\ldots,T$}
    \STATE Generate two views and compute $\mathbf{H}^{(1)},\mathbf{H}^{(2)}$
    \STATE Optimize symmetric active-set InfoNCE
    \STATE Compute detached signals $D_E^{(t)},D_I^{(t)},D_H^{(t)}$

    \IF{$T_b<t\leq T_a$ and $t$ is a $2W$ scheduling epoch}
        \STATE Select $c$ in rotation $H\!\rightarrow\!I\!\rightarrow\!E$
        \STATE Compute $D_{c,\mathrm{prev}},D_{c,\mathrm{cur}}$
        \IF{$D_{c,\mathrm{cur}}\geq\gamma D_{c,\mathrm{prev}}$}
            \STATE $\eta_c\leftarrow\min(1,\eta_c+\Delta\eta)$;
            update $B_c$ and active set
        \ENDIF
    \ENDIF

    \IF{$t$ is a $2W$ reassessment epoch}
        \IF{$t>T_b$}
            \STATE Apply RCR to positive-like reservoir candidates
        \ENDIF
        \STATE Re-rank E/I/H pools and reconcile active sets
    \ENDIF

    \IF{$t>T_b$ and $t$ is a diversity-refresh epoch}
        \STATE Refresh $20\%$ of active candidates within each category
    \ENDIF
\ENDFOR

\RETURN $f_{\theta}$

\end{algorithmic}
\end{algorithm}

\section{Experimental Evaluation}
\label{sec:experiments}
This section evaluates AdNGCL through node classification, hyperparameter sensitivity, component analysis, scheduler and reliability diagnostics, and computational analysis. All methods are compared under a common downstream evaluation protocol across ten benchmark datasets. We first examine overall representation quality and sensitivity to the principal HANS parameters, then isolate the contribution of the main AdNGCL components. We further analyze the behavior of the response-driven scheduler and the candidates identified by the reliability-aware refresh mechanism before examining computational cost and resource usage.

\subsection{Experimental Setup}
\label{subsec:experimental_setup}

\paragraph{Datasets.}
We evaluate AdNGCL on ten graph benchmark datasets spanning citation, co-purchase, co-authorship, Wikipedia, and actor co-occurrence networks. Cora, CiteSeer, PubMed, and DBLP are citation networks; WikiCS is a Wikipedia hyperlink network~\cite{WikiCSdataset}; Computers and Photo are Amazon product co-purchase networks~\cite{AmazonDatasets}; Coauthor-CS and Coauthor-Physics are academic co-authorship networks; and Actor is an actor co-occurrence network constructed from Wikipedia pages~\cite{ActorDataset}. 
Their statistics are summarized in Table~\ref{tab:dataset_hyperparams}. Collectively, these benchmarks span substantial variation in graph size, feature dimensionality, class cardinality, connectivity structure, and application domain, providing a broad test bed for assessing node-classification performance across diverse graph settings.

\begin{table*}[!t]
\caption{Dataset statistics and dataset-specific conventional GCL hyperparameters used for AdNGCL.
``Feat.'' denotes the input feature dimension, while $D_h$ and $D_p$ denote the encoder hidden and projection dimensions, respectively.
$\mathrm{lr}$, $\mathrm{wd}$, and $\tau$ denote the learning rate, weight decay, and contrastive temperature.
$p_e^{(1)}$ and $p_e^{(2)}$ are the edge-drop probabilities for the two views, and
$p_f^{(1)}$ and $p_f^{(2)}$ are the corresponding feature-drop probabilities.
Coauthor-Phy denotes Coauthor-Physics.}

\label{tab:dataset_hyperparams}
\centering
\small
\setlength{\tabcolsep}{2.2pt}

\begin{tabular}{lrrrrccccccccc}
\toprule
&
\multicolumn{4}{c}{\textbf{Dataset Statistics}} &
\multicolumn{9}{c}{\textbf{AdNGCL Hyperparameters}} \\
\cmidrule(lr){2-5}
\cmidrule(lr){6-14}

\textbf{Dataset} &\textbf{Nodes} &\textbf{Edges} &\textbf{Feat.} &\textbf{Class} &$\mathbf{D_h}$ &$\mathbf{D_p}$ &$\mathbf{lr}$ &$\mathbf{wd}$ &$\boldsymbol{\tau}$ &$\mathbf{p_e^{(1)}}$ &$\mathbf{p_e^{(2)}}$ &$\mathbf{p_f^{(1)}}$ &$\mathbf{p_f^{(2)}}$ \\
\midrule

Cora& 2,708 & 10,556 & 1,433 & 7& 256 & 128 & $3{\times}10^{-4}$ & $0$
& 0.5 & 0.1 & 0.3 & 0.3 & 0.2 \\

CiteSeer& 3,327 & 9,104 & 3,703 & 6& 512 & 256 & $1{\times}10^{-4}$ & $1{\times}10^{-3}$& 0.9 & 0.2 & 0.4 & 0.6 & 0.5 \\

Actor& 7,600 & 30,019 & 932 & 5& 256 & 128 & $1{\times}10^{-3}$ & $1{\times}10^{-3}$& 0.7 & 0.2 & 0.6 & 0.1 & 0.3 \\

DBLP& 17,716 & 105,734 & 1,639 & 4& 256 & 128 & $1{\times}10^{-4}$ & $1{\times}10^{-4}$& 0.9 & 0.4 & 0.6 & 0.6 & 0.3 \\

PubMed& 19,717 & 88,648 & 500 & 3& 512 & 256 & $1{\times}10^{-3}$ & $1{\times}10^{-6}$& 0.2 & 0.5 & 0.5 & 0.3 & 0.3 \\

WikiCS& 11,701 & 431,726 & 300 & 10& 384 & 128 & $3{\times}10^{-4}$ & $1{\times}10^{-6}$& 0.2 & 0.3 & 0.6 & 0.1 & 0.2 \\

Computers& 13,752 & 491,722 & 767 & 10& 512 & 256 & $3{\times}10^{-4}$ & $1{\times}10^{-6}$& 0.8 & 0.6 & 0.6 & 0.1 & 0.3 \\

Photo& 7,650 & 238,162 & 745 & 8& 512 & 128 & $1{\times}10^{-3}$ & $1{\times}10^{-6}$& 0.5 & 0.5 & 0.3 & 0.1 & 0.1 \\

Coauthor-CS& 18,333 & 163,788 & 6,805 & 15& 256 & 128 & $5{\times}10^{-4}$ & $1{\times}10^{-4}$& 0.4 & 0.3 & 0.3 & 0.5 & 0.2 \\

Coauthor-Phy& 34,493 & 495,924 & 8,415 & 5& 256 & 128 & $1{\times}10^{-3}$ & $1{\times}10^{-6}$& 0.9 & 0.6 & 0.2 & 0.5 & 0.5 \\

\bottomrule
\end{tabular}
\end{table*}

\paragraph{Baselines.}
We compare AdNGCL with seven representative methods: DGI~\cite{velikovi2019deep}, which maximizes mutual information between local and global representations; GRACE~\cite{DeepGrace2020}, an augmentation-based InfoNCE framework; ProGCL~\cite{xia2022progcl}, which progressively incorporates informative negatives; NegAmplify~\cite{NegAmplify}, which controls the quantity, quality, and diversity of easy, medium, and hard negatives; E2Neg~\cite{E2Neg}, which selects a small set of effective and efficient negatives; IFL-GCL~\cite{IFLGCL}, which addresses false negatives using semantic information derived from InfoNCE; and MHGCL-HA~\cite{MHGCLHA}, which employs multi-head representations and hop augmentation. The latter four were published in 2025-2026, providing comparison with recent developments in GCL.

\paragraph{Evaluation Protocol.}
All node-classification results are averaged over 10 random seeds using identical 10\%/10\%/80\% train/validation/test splits across methods. Representation learning is fully unsupervised. After training, frozen encoder representations are evaluated using a linear classifier trained for up to 1,000 epochs with Adam ($\mathrm{lr}=0.01$, zero weight decay). Validation Micro-F1 is evaluated every 20 epochs, and the checkpoint with the highest validation score is used for test evaluation. Results are reported as mean $\pm$ standard deviation.

\paragraph{Implementation Details.}
For AdNGCL, the conventional GCL hyperparameters are selected using validation performance and reported in Table~\ref{tab:dataset_hyperparams}. Representation learning is performed for 1,000 epochs. The negative-management configuration is shared across datasets: $K=256$, $\gamma=0.99$, $\eta_0=0.05$, and $W=10$. The first 20\% of training forms the bootstrap stage, HANS adapts exposure from 20\% to 80\%, and scheduling opportunities occur every $2W=20$ epochs, yielding 30 opportunities in the adaptive interval under the H$\rightarrow$I$\rightarrow$E rotation. RCR is applied at hardness-reassessment events after bootstrap, while 20\% of active candidates within each hardness category are refreshed every five scheduling intervals. Sensitivity to $K$, $\gamma$, and $W$ is analyzed in Section~\ref{subsec:sensitivity}.

For the baselines, released method- and dataset-specific hyperparameters are used whenever available. When a required hyperparameter is not specified, a common value is used consistently across methods wherever applicable. Each baseline otherwise retains its original architecture, objective, preprocessing, and method-specific operations. All methods use the same data splits and downstream evaluation protocol. For the controlled computational analysis in Section~\ref{subsec:efficiency}, all methods additionally use the same two-layer GCN encoder with 256-dimensional hidden and output representations, training length, splits, seeds, and RTX~4090 24\,GB hardware while retaining their method-specific components.

\subsection{Overall Performance}
\label{subsec}

Table~\ref{tab:RobuResults} reports node-classification Micro-F1 over ten benchmark datasets. AdNGCL achieves the highest mean performance on eight datasets: Cora, CiteSeer, DBLP, PubMed, WikiCS, Photo, Computers, and Coauthor-Physics, and ranks second on Actor. On Coauthor-CS, it remains competitive with the strongest methods.

The gains persist across benchmarks with markedly different scales and structural characteristics. The evaluated graphs range from approximately $2.7$K to $34.5$K nodes, $9.1$K to $495.9$K edges, and feature dimensions from $300$ to $8{,}415$, with substantial variation in connectivity and class cardinality. Within this range, AdNGCL achieves the highest mean Micro-F1 on eight of the ten datasets, including both smaller graphs such as Cora ($84.32\pm0.71$) and substantially larger graphs such as DBLP ($84.07\pm0.33$), PubMed ($86.61\pm0.19$), Computers ($89.11\pm0.24$), and Coauthor-Physics ($95.40\pm0.12$). This consistency across graph size, edge scale, and feature dimensionality suggests that the benefit of adaptive negative exposure is not confined to a particular graph scale or representation regime.

The comparison with negative-oriented GCL methods helps isolate the effect of \emph{how negative information is controlled during training}. ProGCL~\cite{xia2022progcl} progressively incorporates informative negatives, NegAmplify~\cite{NegAmplify} adjusts the sampling ratio among easy, medium, and hard pools, and E2Neg~\cite{E2Neg} reduces the negative set to a compact collection of representative samples. AdNGCL acts on a different control dimension: the candidate reservoir remains bounded, while the number of active negatives in each hardness category is increased only when the corresponding training-response signal satisfies the HANS gate. This exposure control is further coupled with periodic hardness reassessment, reliability-aware replacement, and within-category diversity refresh. The resulting gains across graphs with widely varying node counts, edge scales, and feature dimensions suggest that controlling the \emph{timing and cardinality of negative exposure}, in addition to negative selection itself, provides useful information for contrastive optimization. IFL-GCL~\cite{IFLGCL} addresses the complementary problem of semantic false negatives, while MHGCL-HA~\cite{MHGCLHA} modifies representation construction through multi-head and hop-based augmentation, making their inclusion useful for testing AdNGCL against methods operating on different parts of the GCL pipeline.

The advantage is not uniform across all benchmarks. On Actor, E2Neg obtains the highest mean Micro-F1 of $28.36\pm0.74$, with AdNGCL close behind at $28.10\pm0.59$. On Coauthor-CS, MHGCL-HA reaches $93.09\pm0.31$, compared with $92.56\pm0.12$ for AdNGCL. In contrast, AdNGCL achieves the highest result on the larger Coauthor-Physics graph at $95.40\pm0.12$. These exceptions indicate that no single method is uniformly optimal across all graph configurations, while the overall results show that response-driven exposure remains effective across a broad range of graph sizes, connectivity levels, and feature dimensions.

\begin{table*}[!t]
\caption{Node classification Micro-F1 (mean $\pm$ standard deviation) on benchmark datasets. The highest results are in \textbf{bold}, and the second highest are in \textnormal{\textit{italics}}.}
\label{tab:RobuResults}
\centering
\setlength{\tabcolsep}{4pt}

\begin{tabular}{lcccccccc}
\toprule
\textbf{Dataset}&\textbf{DGI}&\textbf{GRACE}&\textbf{ProGCL}&\textbf{NegAmplify}&\textbf{E2Neg}&\textbf{MHGCL-HA}&\textbf{IFL-GCL}&\textbf{AdNGCL}\\
\midrule
Cora&\textit{83.02$\pm$0.68}&82.59$\pm$1.06&79.20$\pm$0.63&79.83$\pm$1.76&77.45$\pm$1.43&82.63$\pm$0.97&81.87$\pm$0.99&\textbf{84.32$\pm$0.71}\\
CiteSeer&\textit{71.31$\pm$0.57}&67.36$\pm$1.23&67.75$\pm$0.94&70.14$\pm$0.92&68.91$\pm$0.84&70.68$\pm$0.65&68.01$\pm$1.67&\textbf{72.47$\pm$1.00}\\
Actor&26.60$\pm$0.42&27.91$\pm$0.48&26.20$\pm$0.79&27.15$\pm$1.02&\textbf{28.36$\pm$0.74}&27.04$\pm$1.18&26.85$\pm$1.18&\textit{28.10$\pm$0.59}\\
DBLP&82.82$\pm$0.18&{83.60$\pm$0.24}&\textit{83.63$\pm$0.24}&83.31$\pm$0.41&81.46$\pm$0.41&83.45$\pm$0.23&82.32$\pm$1.09&\textbf{84.07$\pm$0.33}\\
PubMed&\textit{85.78$\pm$0.16}&85.46$\pm$0.26&85.09$\pm$0.43&85.01$\pm$0.24&85.72$\pm$0.27&85.77$\pm$0.25&83.49$\pm$1.06&\textbf{86.61$\pm$0.19}\\

\midrule
WikiCS&77.04$\pm$0.94&\textit{79.89$\pm$0.41}&76.63$\pm$1.06&78.34$\pm$0.40&79.49$\pm$0.28&75.12$\pm$0.73&78.47$\pm$0.99&\textbf{80.07$\pm$0.44}\\
Photo&89.52$\pm$0.71&\textit{92.58$\pm$0.38}&91.45$\pm$0.83&91.75$\pm$0.41&89.78$\pm$0.74&92.23$\pm$0.68&90.02$\pm$1.57&\textbf{92.78$\pm$0.32}\\
Computers&87.16$\pm$1.04&88.41$\pm$0.28&85.74$\pm$0.50&\textit{88.43$\pm$0.26}&87.46$\pm$0.86&87.74$\pm$0.80&82.28$\pm$0.48&\textbf{89.11$\pm$0.24}\\
Coauthor-CS&\textit{92.83$\pm$0.12}&92.49$\pm$0.09&92.43$\pm$0.23&91.57$\pm$0.35&92.38$\pm$0.13&\textbf{93.09$\pm$0.31}&91.37$\pm$0.50&92.56$\pm$0.12\\
Coauthor-Physics&94.18$\pm$0.26&95.09$\pm$0.11&95.04$\pm$0.16&\textit{95.14$\pm$0.28}&94.75$\pm$0.13&{95.12$\pm$0.13}&94.59$\pm$0.06&\textbf{95.40$\pm$0.12}\\
\bottomrule
\end{tabular}
\end{table*}

\subsection{Hyperparameter Sensitivity}
\label{subsec:sensitivity}

\begin{figure}[!t]
    \centering
    \subfloat[Gate threshold $\gamma$]{
        \includegraphics[width=.24\textwidth]{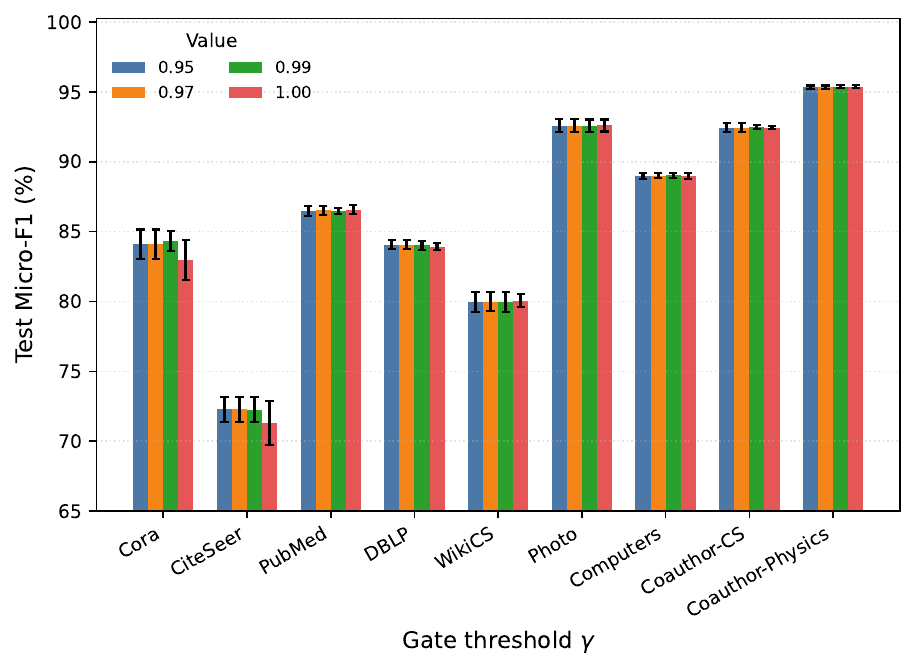}
    }
    \subfloat[Signal-window length $W$]{
        \includegraphics[width=.24\textwidth]{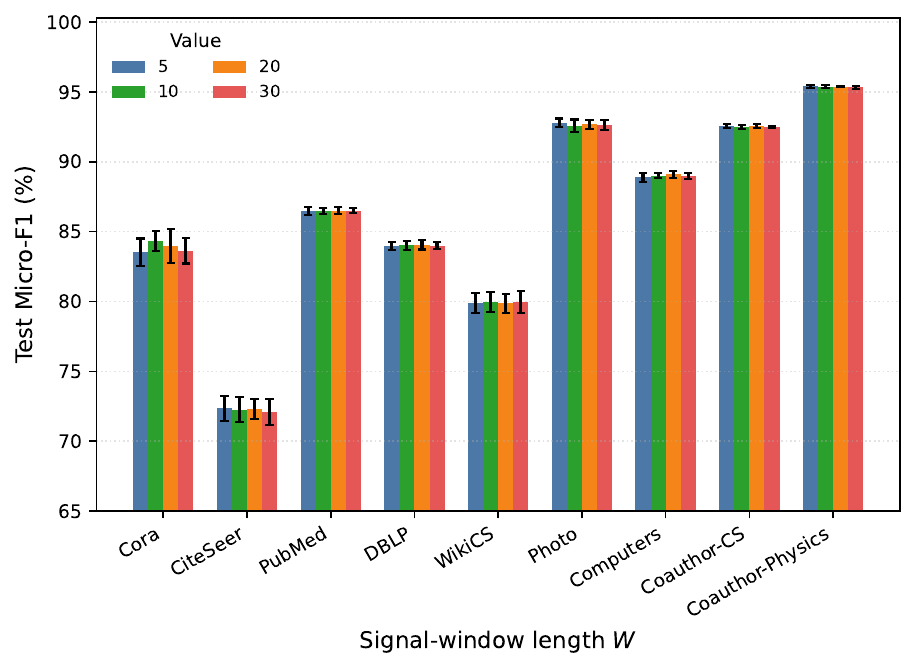}
    }
    \\
    \subfloat[Candidate capacity $K$]{
        \includegraphics[width=.24\textwidth]{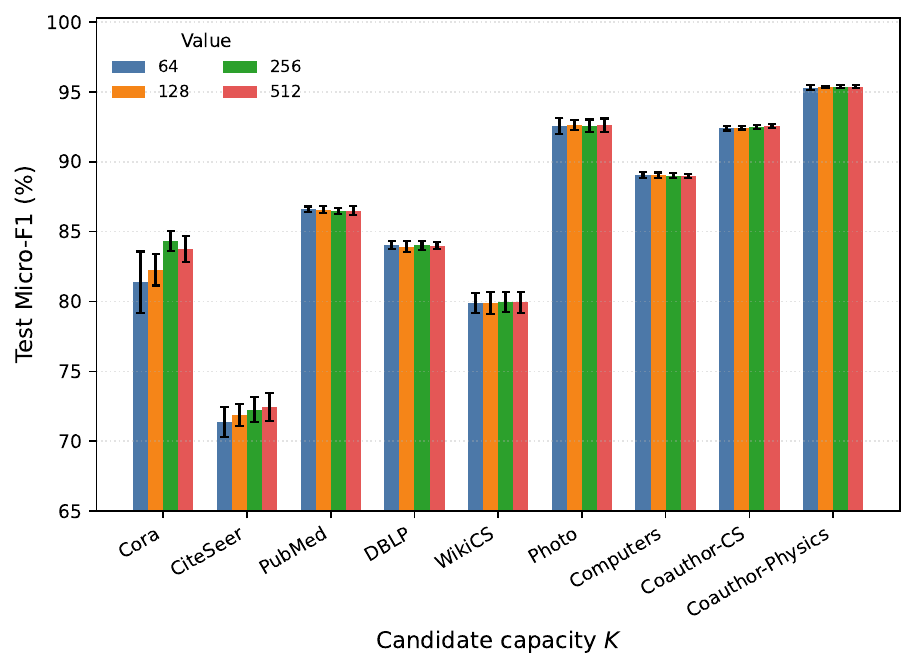}
    }
    \caption{Sensitivity of AdNGCL to
    (a) the HANS gate threshold $\gamma$,
    (b) signal-window length $W$, and
    (c) candidate capacity $K$.
    Performance is generally stable across $\gamma$ and $W$, while $K$ shows somewhat greater dataset-dependent variation and additionally controls the computational and memory requirements of the bounded candidate reservoir.}
    \label{fig:sensitivity}
\end{figure}

We examine the sensitivity of AdNGCL to the three principal HANS control parameters: gate threshold $\gamma$, signal-window length $W$, and candidate capacity $K$. Each parameter is varied independently while the remaining settings are fixed. Sensitivity experiments are conducted on all ten datasets. Actor is omitted from Fig.~\ref{fig:sensitivity} and from the aggregate statistics reported below because its substantially lower Micro-F1 scale differs markedly from the remaining datasets.

\textit{Gate threshold $\gamma$.}
For $\gamma\in\{0.95,0.97,0.99,1.00\}$, the average Micro-F1 varies only from $86.029$ to $86.275$. Relative to the shared setting $\gamma=0.99$, the mean gap to the best dataset-specific value is $0.038$ percentage points, with a maximum of $0.130$. Thus, HANS is generally insensitive to moderate changes in the gate threshold, indicating that its exposure decisions do not require precise dataset-specific calibration.

\textit{Signal window $W$.}
For $W\in\{5,10,20,30\}$, the average Micro-F1 ranges from $86.188$ to $86.275$. The mean gap between the shared setting $W=10$ and the best dataset-specific value is $0.063$ percentage points, with a maximum of $0.212$. Since $W$ controls the temporal window over which the training-response signal is aggregated, this stability indicates that HANS is robust to the exact temporal scale used for scheduling decisions.

\textit{Candidate capacity $K$.}
For $K\in\{64,128,256,512\}$, the average Micro-F1 ranges from $85.845$ to $86.275$. The shared setting $K=256$ is on average only $0.060$ percentage points below the best dataset-specific capacity, with a maximum gap of $0.204$. Variation is somewhat larger than for $\gamma$ and $W$; Cora shows the largest full-range difference of $2.939$ percentage points. Importantly, increasing $K$ enlarges the candidate reservoir and therefore increases the number of negative similarities that must be stored and evaluated, leading to higher memory and computational cost. We therefore use $K=256$ as a practical shared setting: it retains near-best predictive performance across datasets while avoiding the additional resource cost associated with a larger reservoir such as $K=512$.

Overall, AdNGCL remains stable across a broad range of $\gamma$, $W$, and $K$, indicating that HANS does not introduce a substantial dataset-specific tuning burden. Among these parameters, $K$ additionally governs the performance-resource trade-off through the size of the candidate reservoir, which is examined further in Section~\ref{subsec:efficiency}. Having established the robustness of the main control parameters, we next isolate the contribution of the individual AdNGCL mechanisms through the component analysis in Section~\ref{subsec:component_analysis}.

\subsection{Component Analysis}
\label{subsec:component_analysis}

Table~\ref{tab:ablation} isolates the contribution of the principal AdNGCL mechanisms by modifying one component at a time while retaining the remaining training and evaluation settings. The complete model achieves the highest mean Micro-F1 among its variants on six of the ten datasets. More importantly, every ablated variant produces a lower ten-dataset average than the complete model, indicating that the components provide complementary benefits even though their effects are not uniform across individual graphs.

\textit{Training-response-driven exposure.}
The \emph{No Gate} and \emph{All from Start} variants directly examine the central HANS mechanism. Removing the training-response gate lowers the average Micro-F1 by approximately $0.13$ percentage points, with AdNGCL performing better on eight of ten datasets; the largest reductions occur on CiteSeer ($72.47\rightarrow71.96$) and Cora ($84.32\rightarrow83.94$). The \emph{All from Start} variant activates the full candidate reservoir from the beginning of training and produces a larger average reduction of approximately $0.17$ points, with AdNGCL outperforming it on nine datasets. These two ablations support the distinction between \emph{candidate availability} and \emph{active negative exposure}: maintaining a bounded reservoir alone is insufficient, and controlling when additional negatives participate in optimization contributes to the final representation quality.

\textit{Reliability-aware candidate refresh.}
Removing RCR decreases the mean performance on nine of ten datasets, with an average reduction of approximately $0.18$ percentage points. The largest effect occurs on PubMed ($86.61\rightarrow85.95$), followed by Photo ($92.78\rightarrow92.47$) and CiteSeer ($72.47\rightarrow72.17$). RCR targets candidates that appear positive-like in both contrastive directions and replaces them only when reliability and hardness-preservation constraints are satisfied. Its effect therefore complements HANS exposure control by providing reliability-oriented candidate maintenance without deliberately shifting the negative set toward an easier hardness regime. Actor is the only exception, where removing RCR results in a marginal increase from $28.10$ to $28.15$.

\textit{Hardness modeling.}
The \emph{Static Hardness} variant fixes the initial E/I/H assignments rather than periodically reassessing them. AdNGCL performs better on seven datasets, with an average improvement of approximately $0.15$ points and particularly clear differences on CiteSeer, Cora, and Computers. This supports periodic reassessment as a mechanism for keeping the scheduler aligned with the evolving representation space. Replacing hardness-based E/I/H partitions with \emph{Random Strata} yields an average reduction of approximately $0.11$ points, and AdNGCL performs better on eight datasets. The comparatively modest difference is consistent with the role of E/I/H stratification in AdNGCL: the partition itself is not the principal contribution, but provides a structured basis on which response-driven exposure and candidate maintenance operate.

\textit{Diversity maintenance.}
Removing within-category diversity refresh decreases the mean Micro-F1 on all ten datasets and produces the largest average degradation among the individual ablations, approximately $0.21$ percentage points. The effect is most visible on Cora ($84.32\rightarrow83.78$), CiteSeer ($72.47\rightarrow72.04$), Actor ($28.10\rightarrow27.84$), and Photo ($92.78\rightarrow92.57$). Although swapping is a supporting rather than central mechanism, these results indicate that periodically refreshing candidate identities within the same hardness category helps prevent the active negative set from becoming overly static during training.

Overall, the ablations support the proposed components at different levels. The training-response gate and progressive exposure support the central HANS principle of adapting negative participation during optimization; RCR provides complementary reliability control; and dynamic hardness reassessment and diversity refresh maintain the usefulness of the candidate reservoir as representations evolve. The absolute differences of individual ablations are often modest and simplified variants occasionally perform better on particular datasets, so these results are best interpreted as cross-dataset evidence for the integrated framework rather than as evidence that every component is independently optimal for every graph.

\begin{table*}[!t]
\caption{Ablation study of AdNGCL in terms of node classification Micro-F1
(mean $\pm$ standard deviation). Each variant removes or modifies the
corresponding component while keeping the remaining experimental settings
unchanged. The highest result is shown in \textbf{bold}, and the second highest
is shown in \textnormal{\textit{italics}}.}
\label{tab:ablation}
\centering
\setlength{\tabcolsep}{4pt}

\begin{tabular}{lccccccc}
\toprule
\textbf{Dataset}&\textbf{Minus RCR}&\textbf{No Gate}&\textbf{All from Start}&\textbf{Random Strata}&\textbf{Static Hardness}&\textbf{No Swap}&\textbf{AdNGCL}\\
\midrule
Cora&84.31$\pm$0.69&83.94$\pm$0.93&\textit{84.39$\pm$0.84}&\textbf{84.40$\pm$1.06}&83.92$\pm$0.88&83.78$\pm$0.87&84.32$\pm$0.71\\
CiteSeer&72.17$\pm$0.90&71.96$\pm$0.96&72.04$\pm$0.93&\textit{72.30$\pm$0.66}&71.83$\pm$0.75&72.04$\pm$1.18&\textbf{72.47$\pm$1.00}\\
Actor&28.15$\pm$0.59&\textit{28.31$\pm$0.35}&27.90$\pm$0.32&28.24$\pm$0.90&\textbf{28.39$\pm$0.79}&27.84$\pm$0.80&28.10$\pm$0.59\\
DBLP&83.92$\pm$0.34&\textbf{84.13$\pm$0.25}&83.99$\pm$0.30&84.01$\pm$0.32&\textit{84.11$\pm$0.32}&83.98$\pm$0.53&84.07$\pm$0.33\\
PubMed&85.95$\pm$0.23&86.43$\pm$0.30&86.35$\pm$0.26&86.32$\pm$0.26&\textit{86.53$\pm$0.24}&86.35$\pm$0.26&\textbf{86.61$\pm$0.19}\\
\midrule
WikiCS&79.91$\pm$0.72&79.98$\pm$0.67&79.95$\pm$0.72&80.03$\pm$0.80&\textbf{80.11$\pm$0.56}&79.93$\pm$0.73&\textit{80.07$\pm$0.44}\\
Photo&92.47$\pm$0.46&92.67$\pm$0.27&\textit{92.71$\pm$0.34}&\textit{92.71$\pm$0.31}&92.67$\pm$0.31&92.57$\pm$0.51&\textbf{92.78$\pm$0.32}\\
Computers&89.02$\pm$0.19&\textit{89.06$\pm$0.19}&88.90$\pm$0.13&88.86$\pm$0.16&88.80$\pm$0.30&89.01$\pm$0.13&\textbf{89.11$\pm$0.24}\\
Coauthor-CS&92.39$\pm$0.14&92.36$\pm$0.28&92.21$\pm$0.20&92.18$\pm$0.20&92.34$\pm$0.15&\textit{92.54$\pm$0.11}&\textbf{92.56$\pm$0.12}\\
Coauthor-Physics&95.37$\pm$0.11&95.36$\pm$0.11&95.32$\pm$0.09&95.34$\pm$0.07&95.34$\pm$0.07&\textit{95.37$\pm$0.13}&\textbf{95.40$\pm$0.12}\\
\bottomrule
\end{tabular}
\end{table*}

\subsection{Scheduler and Reliability Diagnostics}
\label{subsec:diagnostics}

Table~\ref{tab:diagnostics} examines whether the proposed mechanisms behave as intended during training. Although all datasets receive the same 30 HANS scheduling opportunities, the realized gate-pass rate varies from 57.6\% on WikiCS to 99.0\% on CiteSeer, producing different final easy/intermediate/hard utilization levels. Thus, the scheduling opportunities are predetermined, but the resulting negative exposure is conditioned on the observed category-specific training response rather than increased automatically at every opportunity. The RCR diagnostics show a similarly conservative behavior: only 0.008-1.677\% of reservoir candidates are identified as positive-like, and their same-class proportion exceeds that of the overall reservoir on nine of ten datasets. Actor is the only exception. Together, these results support the intended roles of HANS and RCR: HANS controls active exposure according to training response, while RCR selectively targets a small subset of candidates with stronger positive-like characteristics. \textit{Class labels are used only for this post-hoc diagnostic analysis and never during training.}

\begin{table}[!t]
\caption{Scheduler and reliability diagnostics of AdNGCL. Final utilization is reported as E/I/H. ``SC'' denotes same-class rate measured post hoc, Bank$\rightarrow$Flag compares the same-class rate of the reservoir with that of candidates identified as positive-like. RCR Succ. is the fraction of flagged candidates for which a valid hardness-preserving replacement is found.}
\label{tab:diagnostics}
\centering
\scriptsize
\setlength{\tabcolsep}{2pt}
\resizebox{\columnwidth}{!}{%
\begin{tabular}{lccccc}
\toprule
\textbf{Dataset} &
\textbf{Gate (\%)} &
\textbf{Final E/I/H} &
\textbf{Flag (\%)} &
\textbf{SC: Bank$\rightarrow$Flag (\%)} &
\textbf{RCR Succ. (\%)} \\
\midrule
Cora              & 96.2 & .973/1.000/.919 & .008 & 17.9$\rightarrow$66.3 & 75.0 \\
CiteSeer          & 99.0 & .973/1.000/1.000 & .176 & 17.4$\rightarrow$45.2 & 67.0 \\
Actor             & 79.5 & .851/.837/.729 & 1.522 & 21.3$\rightarrow$20.5 & 17.8 \\
DBLP              & 90.0 & .891/.837/.986 & 1.677 & 32.0$\rightarrow$45.5 & 24.6 \\
PubMed            & 67.6 & .851/.729/.498 & .059 & 35.6$\rightarrow$60.8 & 66.8 \\
WikiCS            & 57.6 & .661/.593/.539 & .062 & 14.3$\rightarrow$46.4 & 70.0 \\
Photo             & 78.6 & .905/.864/.620 & .066 & 16.3$\rightarrow$78.8 & 90.7 \\
Computers         & 71.0 & .688/.661/.824 & .419 & 19.8$\rightarrow$49.2 & 66.8 \\
Coauthor-CS       & 93.3 & 1.000/.932/.878 & .015 & 11.2$\rightarrow$71.1 & 80.7 \\
Coauthor-Physics  & 97.6 & 1.000/1.000/.932 & .112 & 31.9$\rightarrow$81.3 & 76.0 \\
\bottomrule
\end{tabular}%
}
\end{table}

\subsection{Computational Efficiency and Scalability}
\label{subsec:efficiency}

\begin{figure}[!t]
    \centering
    \subfloat[Representation-learning time]{
        \includegraphics[width=.48\textwidth]{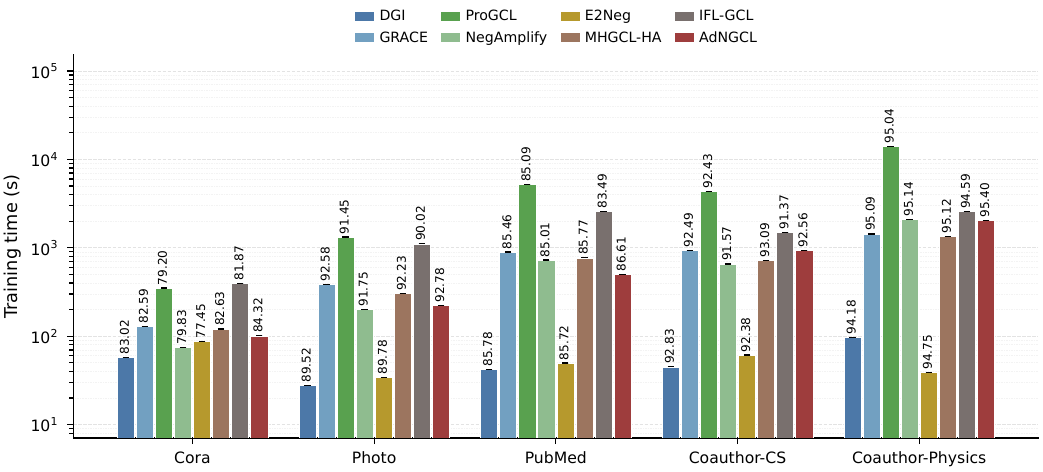}
    }\\
    \subfloat[Peak allocated GPU memory]{
        \includegraphics[width=.48\textwidth]{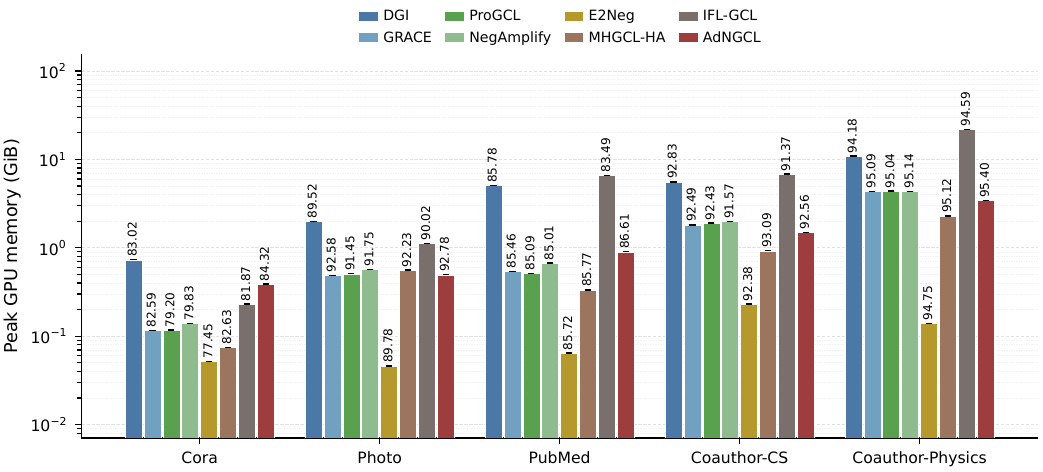}
    }
   \caption{Computational efficiency under the controlled-backbone setting.
    All methods use the same two-layer GCN encoder with 256-dimensional hidden
    and output representations, number of training epochs, data splits, seeds,
    and hardware. (a) Representation-learning time and
    (b) peak allocated GPU memory on five benchmark datasets.
    Micro-F1 values are annotated on the corresponding bars to relate
    computational cost to downstream representation quality.}
    \label{fig:resource_analysis}
\end{figure}

We evaluate computational efficiency under a controlled setting in which all methods use the same two-layer GCN encoder with 256-dimensional hidden and output representations, the same number of training epochs, data splits, seeds, and hardware. Fig.~\ref{fig:resource_analysis} reports representation-learning time and peak allocated GPU memory on five datasets of increasing scale. Micro-F1 scores are annotated on the bars to relate computational cost to representation quality.

\textit{Complexity analysis.}
Let $N$ denote the number of nodes, $d$ the representation dimension, $K$ the per-anchor candidate capacity, and $B^{(t)}\leq K$ the number of active negatives at training step $t$. Excluding the common encoder computation, ordinary active-negative processing requires
\[
\mathcal{O}\!\left(NB^{(t)}d\right),
\]
with a worst-case cost of $\mathcal{O}(NKd)$ when the full reservoir is active. At periodic reassessment events, recomputing candidate similarities and reranking the reservoir into hardness categories requires $\mathcal{O}(NKd+NK\log K)$. RCR additionally evaluates a temporary $K$-candidate proposal set for each anchor; computing proposal similarities requires $\mathcal{O}(NKd)$, while eligibility checking and hardness-preserving replacement selection introduce a worst-case $\mathcal{O}(NK^2)$ term. A complete reliability and hardness refresh therefore has worst-case complexity
\[
\mathcal{O}\!\left(NKd+NK\log K+NK^2\right),
\]
but this cost is incurred only at scheduled reassessment events rather than at every optimization step. Persistent candidate indices require $\mathcal{O}(NK)$ storage and node representations require $\mathcal{O}(Nd)$ memory; the temporary RCR proposal set also requires $\mathcal{O}(NK)$ indices without increasing the persistent reservoir capacity. Thus, for fixed $K$, the negative-management cost remains linear in $N$, while increasing $K$ enlarges both the candidate search space and resource requirements, including the worst-case quadratic replacement-matching term during RCR.

\textit{Empirical analysis.}
AdNGCL incurs additional computation from training-response scheduling, periodic hardness reassessment, and candidate maintenance, and therefore is not designed to minimize training time alone. Nevertheless, its runtime remains competitive with several GCL baselines. On PubMed, AdNGCL requires $497.86$\,s, compared with $727.02$\,s for NegAmplify, $892.74$\,s for GRACE, $2574.95$\,s for IFL-GCL, and $5179.09$\,s for ProGCL. On Coauthor-Physics, it requires $2032.40$\,s, comparable to NegAmplify ($2090.15$\,s) and below IFL-GCL ($2559.84$\,s) and ProGCL ($14069.39$\,s), while achieving the highest Micro-F1.

The memory behavior becomes more favorable as graph scale increases. AdNGCL uses $1.50$\,GiB on Coauthor-CS, compared with $5.57$\,GiB for DGI, $1.82$\,GiB for GRACE, $1.90$\,GiB for ProGCL, $1.97$\,GiB for NegAmplify, and $6.82$\,GiB for IFL-GCL. On the larger Coauthor-Physics graph, it requires $3.41$\,GiB, remaining below DGI ($10.88$\,GiB), GRACE ($4.37$\,GiB), ProGCL ($4.39$\,GiB), NegAmplify ($4.32$\,GiB), and IFL-GCL ($21.74$\,GiB). This behavior is consistent with the bounded-reservoir design: negative management is restricted by the fixed candidate capacity, while only the currently active subset participates in contrastive optimization. The scheduler and refresh mechanisms therefore add computation without requiring unrestricted growth of the negative set.

AdNGCL does not dominate every resource metric; lightweight approaches such as E2Neg, and in some cases MHGCL-HA, require less memory or training time. Rather, the results demonstrate a practical accuracy-resource trade-off: AdNGCL maintains strong representation quality while keeping peak memory moderate on larger graphs. 

\section{Conclusion}
\label{sec:conclusion}

This paper introduced AdNGCL, a graph contrastive learning framework that separates the negative candidates available to an anchor from the subset actively exposed during optimization. Its central component, Hardness-Aware Negative Scheduling (HANS), controls the growth of easy, intermediate, and hard negative exposure using recent category-specific training response rather than increasing participation automatically at fixed training intervals. Dynamic hardness reassessment keeps the candidate organization aligned with the evolving representation space, while reliability-aware candidate refresh and within-category diversity refresh provide complementary candidate maintenance without changing the standard contrastive objective. Across ten benchmark datasets, AdNGCL achieves the highest mean Micro-F1 among the evaluated methods on eight datasets, ranks second on Actor, and remains competitive on Coauthor-CS. Component ablations show that response-gated exposure, reliability maintenance, hardness reassessment, and diversity refresh provide complementary benefits, while the scheduler diagnostics confirm that identical scheduling opportunities can produce different realized exposure levels across datasets. The computational analysis further shows that the bounded candidate reservoir provides a controllable accuracy-resource trade-off, although periodic reassessment and candidate maintenance introduce additional training cost. Overall, the results support adaptive negative exposure as a useful control dimension in graph contrastive learning beyond negative selection alone. Future work can investigate more flexible exposure policies and lower-cost candidate maintenance for larger-scale graphs.

\bibliographystyle{IEEEtran}
\bibliography{references}

\end{document}